\documentclass[11pt]{article}

\usepackage[preprint]{acl}

\usepackage{times}
\usepackage{latexsym}

\usepackage[T1]{fontenc}
\usepackage[utf8]{inputenc}

\usepackage{microtype}

\usepackage{inconsolata}

\usepackage{graphicx}
\usepackage{booktabs}
\usepackage{CJKutf8}

\usepackage{longtable}
\usepackage{array}
\usepackage{tabularx}
\usepackage{enumitem}

\title{LawCompass: Navigating from Legal QA to Multi-Agent Deep Research with Grounded Evidence}

\author{
 \textbf{Xiaoxia Cheng\textsuperscript{1}},
 \textbf{Linnan Wang\textsuperscript{1}},
 \textbf{Jiahao Ma\textsuperscript{1}},
 \textbf{Zhichuan Ye\textsuperscript{1}},
 \textbf{Xuemei Zhou\textsuperscript{1}},
 \\
 \textbf{Chuanyu Tong\textsuperscript{2}},
 \textbf{Bo Jiang \textsuperscript{1}\thanks{Corresponding author.}},
 \textbf{Qing Zhu\textsuperscript{1}\footnotemark[1]},
\\
 \textsuperscript{1}Anhui University,
 \textsuperscript{2}Tsinghua University,
\\
 \small{
   zjucxx@zju.edu.cn, e125211014@stu.ahu.edu.cn, \{jiangbo, zhuqing\}@ahu.edu.cn
 }
}

\usepackage[most]{tcolorbox}

\newtcolorbox{promptbox}{
  breakable,
  colback=gray!4,
  colframe=gray!55,
  boxrule=0.4pt,
  arc=1mm,
  left=2mm,
  right=2mm,
  top=1.5mm,
  bottom=1.5mm,
  before skip=6pt,
  after skip=8pt,
  fontupper=\small
}

\begin{document}
\begin{CJK*}{UTF8}{gbsn}
\maketitle
\begin{abstract}
Recent advances in Large Language Models (LLMs) and Retrieval-Augmented Generation (RAG) have significantly democratized access to legal information. Nevertheless, most existing legal assistants remain confined to multi-turn conversational QA, failing to support complex legal tasks that require systematic evidence retrieval, multi-step reasoning, and report-level synthesis. In this paper, we present \textbf{LawCompass}, an evidence-grounded legal assistant that navigates the transition from standard Legal QA to multi-agent deep research. LawCompass provides three task-oriented functions: \textbf{Legal QA}, which delivers precise, evidence-backed answers to legal questions; \textbf{Professional Retrieval}, which enables structured exploration of statutes and judicial cases via query rewriting; and \textbf{Deep Research}, which employs a multi-agent workflow to decompose complex legal tasks and synthesize comprehensive research reports. Crucially, LawCompass maintains explicit citation links across all modules, empowering users to directly verify system outputs against original legal sources. Evaluation results demonstrate that LawCompass provides a practical and scalable paradigm for transforming conversational AI into trustworthy and evidence-grounded legal research assistance.

\end{abstract}

\section{Introduction}
Recent advances in Large Language Models (LLMs) \cite{openai2024gpt4technicalreport, grattafiori2024llama3herdmodels, deepseekai2026deepseekv4} have significantly improved the capabilities of AI assistants in complex knowledge-intensive tasks \cite{lai2023largelanguagemodelslaw, li2024largelanguagemodelsfinance}.
In the legal domain, LLMs have demonstrated promising performance in legal question answering \cite{craciun-etal-2025-graf, t-y-s-s-etal-2025-cocolex} and legal document understanding \cite{wei-etal-2025-legalcore, belfathi-etal-2025-simple}.
However, legal practice demands not only fluent answers but also verifiable and trustworthy evidence.

\begin{figure}[t]
  \includegraphics[width=\columnwidth]{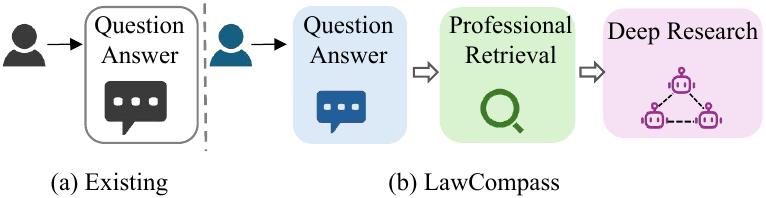}
  \caption{Comparison between existing legal question answering systems and our proposed LawCompass. (a) Existing systems focus on multi-turn question answering; (b) Our proposed LawCompass extends this by conducting multi-agent deep research.}
  \label{fig:introduction}
\end{figure}

Recently, retrieval-augmented generation (RAG) \cite{NEURIPS2020_6b493230} offers a promising path toward more grounded legal assistance by combining the language understanding capabilities of LLMs with external legal corpora \cite{hu-etal-2024-ella, luo-etal-2025-automating, li-etal-2025-unilr, Cui_2026}. 
For instance, ELLA \cite{hu-etal-2024-ella} leverages RAG strategies to deliver interpretable and informative legal advice, while MASER \cite{shengbinyue-etal-2025-multi} enhances interactive consultation via multi-agent simulation of lawyer–client dialogues.
While these approaches \cite{hu-etal-2024-ella, shengbinyue-etal-2025-multi, Cui_2026} improve factual grounding by incorporating external legal resources, existing systems are often confined to multi-turn conversational QA, as shown in Figure \ref{fig:introduction} (a).
They rarely support complex legal tasks that require systematic evidence retrieval, multi-step reasoning, and report-level synthesis.
In practice, legal information seeking is inherently progressive. 
A practitioner typically initiates the workflow with a direct query, transitions to a structured examination of the underlying legal authorities, and ultimately requires an in-depth analysis that cross-references statutes, judicial cases, and empirical evidence.
This process requires not only accurate retrieval but also coordinated reasoning across multiple sources of legal information. 
Therefore, there is a growing need for legal systems that can 
seamlessly bridge the gap between superficial legal QA and comprehensive legal research.

In this paper, we propose \textbf{LawCompass}, a unified evidence-grounded legal system that empowers users to seamlessly navigate from direct legal question answering to comprehensive deep legal research, as shown in Figure \ref{fig:introduction} (b).
Built upon a RAG framework, LawCompass aggregates external knowledge from heterogeneous legal sources, including statutes, judicial cases, and user-uploaded documents, to rigorously ground system outputs across all target applications.
Specifically, LawCompass provides three task-oriented functions. 
First, \textbf{Legal QA} provides concise answers to legal questions, ensuring every claim is backed by explicit legal evidence.
Second, \textbf{Professional Retrieval} enables users to search and explore relevant legal authorities, including statutes and judicial cases. 
To improve retrieval accuracy and result clarity, it incorporates query rewriting and structured result presentation. 
Third, \textbf{Deep Research} extends the retrieval-augmented framework with a multi-agent workflow for complex legal research tasks. 
It decomposes broad research objectives into subtasks, collects and verifies multi-source evidence, analyzes relevant legal authorities, and synthesizes comprehensive research reports. 
Compared with standard Legal QA, Deep Research is better suited for questions that require multi-step reasoning, evidence comparison, and report-level legal analysis.
By integrating structured legal retrieval and multi-agent deep research into a single demonstration system, LawCompass offers a practical interface for supporting diverse legal information needs with traceable evidence.
The user study further demonstrates the usability and practical value of LawCompass, showing that users benefit from its evidence-grounded responses, structured retrieval results, and comprehensive Deep Research reports.

\section{Related Work}
Large language models (LLMs) \cite{openai2024gpt4technicalreport, deepseekai2026deepseekv4, grattafiori2024llama3herdmodels} have shown strong potential for legal information access and reasoning \cite{lai2023largelanguagemodelslaw, wei-etal-2025-legalcore}.
However, legal applications require more than fluent responses, as system outputs should be accurate, interpretable, and grounded in authoritative legal sources. 
To mitigate hallucinations, Retrieval-Augmented Generation (RAG) \cite{NEURIPS2020_6b493230} has been widely adopted.
For example, ELLA \cite{hu-etal-2024-ella} and other recent legal assistants \cite{panchal2025lawpalretrievalaugmented, Cui_2026} integrate evidence-aware designs by fetching relevant articles.
However, these systems remain predominantly confined to  short-form conversational QA. 
To unlock advanced cognitive capabilities, recent works have explored multi-agent frameworks \cite{shengbinyue-etal-2025-multi, deepanshu2026nyayaaiaipoweredlegalassistant}.
Within the legal domain, MASER \cite{shengbinyue-etal-2025-multi} employs a multi-agent framework to simulate lawyer--client dialogues for evaluation and data generation.
Differing from this simulation paradigm, LawCompass utilizes a highly orchestrated multi-agent workflow that moves beyond the concise response format of Legal QA to better support complex legal research tasks.

\section{Framework and Usage Example}

LawCompass adopts a Unified Task Entry interface that routes user requests to three task-oriented modes: \textit{(i)} Legal QA, \textit{(ii)} Professional Retrieval, and \textit{(iii)} Deep Research. 
The three modes share common services for document input, citation rendering, session management, and authenticated access control, while differing in task complexity, retrieval strategy, and output structure.

\begin{figure*}[t]
  \includegraphics[width=\textwidth]{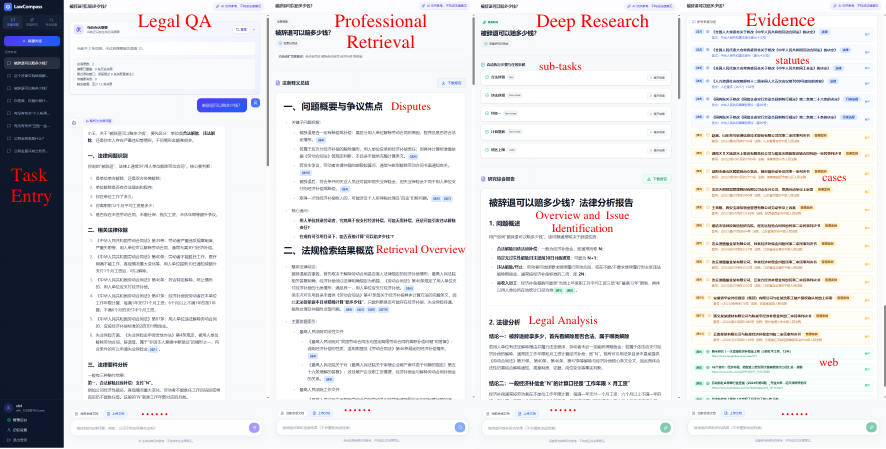}
  \caption{
  Demonstration of the LawCompass interface, including Unified Task Entry, Legal Question Answering, Professional Retrieval, and Deep Research, together with a unified Grounded Evidence view for inspecting supporting legal authorities.
}
  \label{fig:usage}
\end{figure*}

\subsection{Unified Task Entry}
As shown in Figure \ref{fig:usage} (left), LawCompass features an intuitive input interface to serve all three operational modes.
Users can express natural language queries, leverage pre-curated template questions, or ingest local documents to serve as grounded contextual materials. 
This interface supports query-only, document-only, and joint query-document inputs, empowering LawCompass to orchestrate diverse real-world workflows ranging from legal consultation to document-centric deep research.

\subsection{Legal Question Answering}
The Legal Question Answering module provides concise legal answers grounded in retrieved evidence.
Given a legal question, the module retrieves supporting statutes, cases, or uploaded document fragments and generates an answer with inline citation tags. 
These citation tags are clickable, allowing users to open a source panel and inspect the title, article number, validity status, case metadata, original text, or external source link. 
This allows users to move from an answer-level summary to source-level verification.
Furthermore, the module leverages conversational memory to maintain context across multi-turn dialogues. 
This module is optimized  for routine legal consultations that require immediate responses without conducting extensive legal research.

\subsection{Professional Retrieval}
The Professional Retrieval module empowers users to execute structured, in-depth navigation across  heterogeneous legal corpora, including statutes, judicial cases, and supporting legal resources. 
To improve retrieval quality, LawCompass first performs query rewriting and dual-path retrieval, allowing users to discover relevant legal authorities even when the original query is incomplete or expressed in non-legal terminology. 
The retrieved results are then organized into a structured, evidence-oriented report with four sections: \textbf{1) Summary of the Legal Question and Key Disputes}; \textbf{2) Overview of Retrieved Statutory or Case Materials}; \textbf{3) Detailed Analysis of Core Legal Provisions, Adjudication Rules, or Representative Cases}; and \textbf{4) Retrieval Conclusions, Practical Recommendations, and Risk Alerts}.
This structured presentation allows users to conveniently inspect legal sources, compare relevant materials, and collect evidence for deeper legal research.

\subsection{Deep Research}
While Legal QA and Professional Retrieval target highly localized information needs, real-world legal practice frequently requires systematic investigations across heterogeneous source materials and analytical dimensions.
To support such scenarios, LawCompass provides a Multi-Agent Deep Research module, as shown in Figure \ref{fig:usage}.
Upon receiving a research objective, the system initiates a highly orchestrated multi-agent workflow.
The \textit{Planner Agent} first decomposes the objective into a set of actionable subtasks.
Subsequently, specialized \textit{Executor Agents} are dispatched in parallel to harvest, cross-verify, and analyze empirical evidence across statutes, judicial cases, web-scale repositories, and user-uploaded documentation. 
During this process, intermediate findings are presented as subtask reports, allowing users to assess whether the system has covered crucial analytical dimensions.
Once all subtasks have been completed, the \textit{Synthesis Agent} integrates the collected findings into a comprehensive legal research report with six sections: \textbf{1) Overview and Issue Identification}; \textbf{2) Detailed Legal Analysis}; \textbf{3) Statutory Provisions}; \textbf{4) Case References}; \textbf{5) Risk Alerts}; and \textbf{6) Liability Disclaimer}.

\section{System Architecture and Implementation}

\begin{figure*}[t]
  \includegraphics[width=\textwidth]{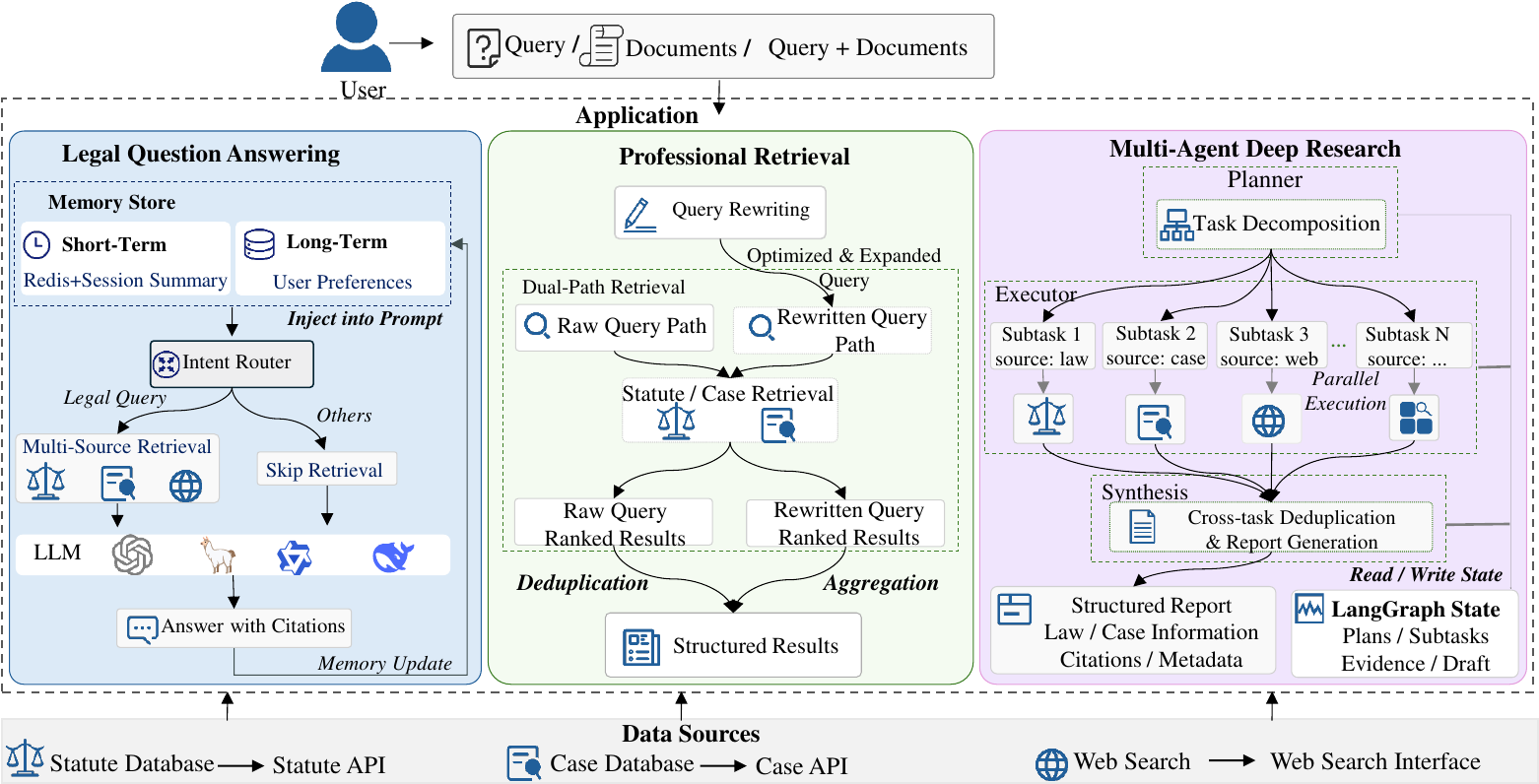}
  \caption{System architecture of LawCompass. The system connects role-based user interactions with three application modules, including legal question answering, professional retrieval, and deep research, while retrieving evidence from statute, case, and web search interfaces to support answer generation and report synthesis.}
  \label{fig:framework}
\end{figure*}

This section presents the  architecture and detailed implementation of LawCompass.
Figure~\ref{fig:framework} provides an overview of the backend components supporting Legal QA, Professional Retrieval, and multi-agent Deep Research.

\subsection{LLM and Data Services}
\paragraph{LLM.} LawCompass employs DeepSeek-v4 \cite{deepseekai2026deepseekv4}
as its primary reasoning engine to orchestrate core workflows, including intent classification, query rewriting, research planning, and multi-source report synthesis. Other LLMs, such as GPT~\cite{openai2026gpt54} and Llama~\cite{grattafiori2024llama3herdmodels}, 
are also supported.

\paragraph{Data Services.}
To support evidence-grounded legal question answering and deep research, we collect  220,000 statutory provisions from Lawyee\footnote{https://data.lawyee.net/} and 100,000 court cases from China Judgements Online\footnote{https://wenshu.court.gov.cn/}, as shown in Figure \ref{fig:framework} (bottom). 
These data are stored in a Milvus vector database.
Furthermore, to bridge potential temporal gaps in static databases, we integrate real-time web search augmentation powered by Tavily\footnote{https://tavily.com/}.

\subsection{Legal Question Answering}
As shown in Figure \ref{fig:framework} (left), the module comprises four key components: a memory store, an intent router, a retrieval engine, and an answer generator.

\paragraph{Memory Store.} LawCompass maintains both short-term and long-term memory, implemented using Redis and MySQL, to support personalized assistance. Short-term memory stores session-level interaction summaries, including key facts, unresolved issues, and previous advice within the current conversation. Long-term memory captures persistent user preferences and historical contexts, such as response style, citation formats, and languages.
\paragraph{Intent Router.}
To avoid introducing unnecessary latency and cost, LawCompass adopts a two-tier routing strategy for user input. The first tier is a heuristic router that uses predefined regular expressions to detect common trivial inputs, including greetings, system-level questions, and overly broad legal keywords. 
These inputs are routed directly to the corresponding response flow without an additional model call. 
The second tier is an LLM-based semantic classifier. 
Inputs not captured by the heuristic router are assigned one of four labels: chitchat, clarify\_legal, assistant\_meta, or legal\_query. 
Only the legal\_query proceeds to retrieval and generation.
\paragraph{Retrieval and Generation.} 
The retrieval results are obtained using the method described in Section~\ref{profession-retrieval}, and are combined with relevant conversational memory for grounded answer generation, as detailed in Section~\ref{generation}.

\subsection{Professional Retrieval} \label{profession-retrieval}
As illustrated in Figure \ref{fig:framework} (middle), this module is operated through query rewriting, dual-path retrieval, result aggregation and structured presentation generation, with the generation process described in Section~\ref{generation}.

\paragraph{Query Rewriting.} 

Legal search queries are often incomplete, ambiguous, or expressed in non-professional language. 
To improve retrieval effectiveness, LawCompass first employs an LLM to reformulate user queries into retrieval-oriented representations. 
The rewritten query enriches legal terminology and expands 3--5 relevant concepts while preserving the original user intent.

\paragraph{Dual-Path Retrieval.}
To balance precision and recall, LawCompass performs retrieval using both the original user query and the rewritten query. 
The original query preserves the user's explicit information needs, while the rewritten query improves coverage of legally relevant concepts and terminology.
Each query is independently executed against multiple legal knowledge sources using both keyword-based and vector-based retrieval.
Vector search is sorted in descending order by cosine similarity, while keyword search is sorted by time, giving preference to more recent results.
The retrieved results are then reorganized according to legal authority: statutory materials are ranked according to their hierarchy of legal authority, and cases are ranked according to their level of reference authority in similar-case analysis, while the original retrieval order is preserved among sources at the same level.

\paragraph{Result Aggregation.} 

Retrieved results from the dual-path retrieval are subsequently merged and re-ranked to construct a unified evidence set. 
The aggregation process first removes duplicate results, merges candidates according to their retrieval scores, and organizes the final results into structured retrieval outputs through the evidence-grounded generation process described in Section \ref{generation}.
The resulting evidence set can be directly provided to the Multi-Agent Deep Research module.

\subsection{Multi-Agent Deep Research}
Unlike Legal QA, which provides concise answers, the deep research module performs end-to-end legal research through planner, executor, and synthesis agents coordinated by LangGraph \cite{langgraph}. 
As shown in Figure~\ref{fig:framework} (right), LangGraph maintains the evolving plan, subtasks, evidence, and draft report state.

\paragraph{Planner.} 
The Planner Agent coordinates the Deep Research workflow by converting a complex legal query into an executable research plan. It identifies the main legal issues and decomposes the query into 3--5 complementary subtasks, each with a distinct focus and specified source scope, such as legal sources, web sources, or both.
The output plan is validated against a predefined schema to ensure the required task count, clear objectives, and complete information for task execution.
The invalid plan is first passed to a structural repair module,
which fills in missing information, removes invalid tasks, and adjusts the task count.
If repair fails, LawCompass applies a deterministic fallback template with three standard subtasks: Legal Authority Compilation, Case and Adjudication Rule Retrieval, and Practical Risk Analysis. This recovery design maintains workflow continuity while preserving the flexibility of customized planning.

\paragraph{Executor.} 
The Executor Agent is responsible for executing the subtasks generated by the Planner Agent through tool-augmented retrieval.
For each subtask, it reads the planned query and search\_type, then routes the request to the appropriate retrieval channels, including statutory databases, judicial case repositories, web search APIs, and user-provided documents.
After retrieval, the collected evidence is organized into a structured context and used to generate a subtask-level summary using a legal research prompt.
The summary presents the key findings, cites relevant statutory provisions and case numbers, and gives priority to sources with higher legal authority or case reference levels when ranking metadata is provided.

\paragraph{Synthesis.} 

The Synthesis Agent functions as the terminal consolidation layer of the Deep Research workflow. 
It collects the subtask-level summaries and source materials produced by the parallel Executor Agents, deduplicates evidence across statutory, case, web, and document sources, and attaches citation labels to the remaining references.
Using the subtask findings and the  remaining references,
the Synthesis Agent generates a structured legal analysis report. The report follows a predefined template covering the overview and issue identification, detailed legal analysis, statutory provisions, case references, risk alerts, and liability disclaimer. 

\subsection{Evidence-Grounded Generation} \label{generation}
This shared component is employed across all modules to transform retrieved legal materials into coherent answers or a report through hierarchy-aware 
source ordering and mandatory citation constraints.

\paragraph{Hierarchy-Aware Source Ordering.}
Unlike general-domain RAG systems that rank context mainly by semantic similarity, LawCompass incorporates the legal authority of retrieved sources.
During subtask summarization and final report synthesis, statutory materials and cases are organized according to their respective authority levels or case reference levels, with higher-ranked sources prioritized in the generated analysis. 
When such ranking metadata is available, the generation prompts explicitly instruct the LLM to present and cite higher-authority materials before lower-authority ones, so that conclusions are grounded in the most authoritative available evidence.

\paragraph{Mandatory Citation.}

LawCompass adopts a structured citation framework to improve output verifiability. Retrieved statutes, cases, web pages, and uploaded documents are assigned numbered source tags, such as \texttt{[statute 1]}, \texttt{[case 2]}, \texttt{[web 1]}, and \texttt{[document 1]}. These sources are represented in a source catalogue with key metadata, including statutory hierarchy, article numbers, case numbers, courts, case reference levels, web links, and document locations.
During generation, the LLM is instructed to cite only tags from the provided source catalogue.
In the user interface, these citation tags are rendered as clickable links, allowing users to inspect the corresponding source metadata and original excerpts.

\section{Evaluation}
We evaluate LawCompass from three perspectives: retrieval quality \S\ref{Retrieval-quality}, user study \S\ref{user-quality}, and a representative case study presented in Appendix~\ref{sec:abridged_legal_qa_example}.

\subsection{Retrieval Quality}\label{Retrieval-quality}
High-quality retrieval is essential for LawCompass, as all three task-oriented functions rely on retrieved legal sources to generate or present evidence-grounded outputs. 
We conduct a human relevance-based retrieval evaluation using a set of 20 legal queries covering different legal information needs. 
For each query, we collect the top-5 evidence passages returned by the system and ask two annotators with legal backgrounds to independently judge the relevance of each passage. 
Each retrieved passage is labeled as relevant, partially relevant, or irrelevant according to whether it provides useful legal authority for addressing the user query. 
Based on these annotations, we report Precision@5, nDCG@5, and the average relevance score. 
The results in Table~\ref{tab:retrieval_quality} confirm the effectiveness of dual-path retrieval in LawCompass.

\begin{table}[h]
\centering
\small
\begin{tabular}{lccc}
\toprule
Method & P@5 $\uparrow$ & nDCG@5 $\uparrow$ & Rel. Score $\uparrow$ \\
\midrule
Original & 76.0 & 85.5 & 1.10 \\
Query-Rewritten & 75.0 & 85.9 & 1.11 \\
LawCompass & 79.0 & 90.9 & 1.28\\
\bottomrule
\end{tabular}
\caption{Retrieval quality evaluation on 20 legal queries. 
}
\label{tab:retrieval_quality}
\end{table}

\subsection{User Study}\label{user-quality}
We further conducted a task-based user study to evaluate the usability and practical value of LawCompass. 
Specifically, we invited participants with and without legal backgrounds to examine whether the system can support both ordinary users and users with professional legal knowledge. 
Each participant was asked to complete three representative tasks using LawCompass. 
After completing the tasks, participants evaluated LawCompass on a five-point Likert scale in terms of usability, usefulness, clarity, evidence-grounded trustworthiness, and the utility of Deep Research.
Table~\ref{tab:user_study} reports the average scores obtained by rating the collected open-ended feedback.
The results indicate that LawCompass is well received by both groups, with particularly strong ratings for usefulness, Deep Research value, and overall satisfaction.

\begin{table}[h]
\centering
\small
\begin{tabular}{lccc}
\toprule
Dimension & Legal & Non-legal & Overall \\
\midrule
Ease of Use & 3.9 & 4.2 & 4.1 \\
Usefulness & 4.0 & 4.5 & 4.3 \\
Evidence Trust & 3.8 & 4.5 & 4.2 \\
Retrieval Clarity & 3.9 & 4.3 & 4.1 \\
Deep Research Value & 4.2 & 4.5 & 4.4 \\
Overall Satisfaction & 4.1 & 4.4 & 4.3 \\
\bottomrule
\end{tabular}
\caption{User study results across participants with and without legal backgrounds. Scores are based on a 5-point Likert scale.}
\label{tab:user_study}
\end{table}

\section{Conclusion}
We present LawCompass, an evidence-grounded legal system that supports Legal QA, Professional Retrieval, and multi-agent Deep Research within a unified interface. By combining heterogeneous legal sources with task-oriented retrieval and a coordinated multi-agent workflow, LawCompass enables users to move from direct legal information seeking to complex, report-level research. The retrieval evaluation, user study, and representative case studies validate the effectiveness of LawCompass in supporting practical legal research tasks.

\section*{Limitations}
This paper presents LawCompass, a unified legal research system that extends conventional legal QA to multi-agent Deep Research.
Despite its practical utility, LawCompass still has several limitations.
First, its retrieval coverage depends on the completeness and update frequency of the local database. Newly issued legal documents, recent cases, and local regulations may not be included promptly. 
Second, although the Deep Research module improves the structure and interpretability of complex questions by decomposing them into multiple subtasks, this multi-stage process also introduces additional latency and computational cost.
\bibliography{custom}

\begin{thebibliography}{19}
\providecommand{\natexlab}[1]{#1}

\bibitem[{Belfathi et~al.(2025)Belfathi, Hernandez, Monceaux, and Dufour}]{belfathi-etal-2025-simple}
Anas Belfathi, Nicolas Hernandez, Laura Monceaux, and Richard Dufour. 2025.
\newblock \href {https://doi.org/10.18653/v1/2025.argmining-1.15} {A simple but effective context retrieval for sequential sentence classification in long legal documents}.
\newblock In \emph{Proceedings of the 12th Argument mining Workshop}, pages 160--167, Vienna, Austria. Association for Computational Linguistics.

\bibitem[{Craciun et~al.(2025)Craciun, Sm{\u{a}}du, Cercel, and Cercel}]{craciun-etal-2025-graf}
Cristian-George Craciun, R{\u{a}}zvan-Alexandru Sm{\u{a}}du, Dumitru-Clementin Cercel, and Mihaela-Claudia Cercel. 2025.
\newblock \href {https://doi.org/10.18653/v1/2025.findings-acl.659} {{GRAF}: Graph retrieval augmented by facts for {R}omanian legal multi-choice question answering}.
\newblock In \emph{Findings of the Association for Computational Linguistics: ACL 2025}, pages 12708--12742, Vienna, Austria. Association for Computational Linguistics.

\bibitem[{Cui et~al.(2026)Cui, Ning, Li, Li, Ya, Chen, Ling, Tian, and Yuan}]{Cui_2026}
Jiaxi Cui, Munan Ning, Zongjian Li, Hao Li, Yang Ya, Bohua Chen, Bin Ling, Yonghong Tian, and Li~Yuan. 2026.
\newblock \href {https://doi.org/10.1016/j.fmre.2026.03.026} {Chatlaw: A multi-agent legal assistant based on a role-aligned mixture-of-experts architecture}.
\newblock \emph{Fundamental Research}.

\bibitem[{Deepanshu et~al.(2026)Deepanshu, Saxena, Rana, Varshney, and Laskar}]{deepanshu2026nyayaaiaipoweredlegalassistant}
Deepanshu, Divi Saxena, Deepali Rana, Ayesha Varshney, and Sahinur~Rahman Laskar. 2026.
\newblock \href {https://arxiv.org/abs/2605.10155} {Nyayaai: An ai-powered legal assistant using multi-agent architecture and retrieval-augmented generation}.
\newblock \emph{Preprint}, arXiv:2605.10155.

\bibitem[{DeepSeek-AI(2026)}]{deepseekai2026deepseekv4}
DeepSeek-AI. 2026.
\newblock Deepseek-v4: Towards highly efficient million-token context intelligence.

\bibitem[{Grattafiori et~al.(2024)Grattafiori, Dubey, Jauhri, Pandey, Kadian, Al-Dahle, Letman, Mathur, Schelten, Vaughan, Yang, Fan, Goyal, Hartshorn, Yang, Mitra, Sravankumar, Korenev, Hinsvark, Rao, Zhang, Rodriguez, Gregerson, Spataru, Roziere, Biron, Tang, Chern, Caucheteux, Nayak, Bi, Marra, McConnell, Keller, Touret, Wu, Wong, Ferrer, Nikolaidis, Allonsius, Song, Pintz, Livshits, Wyatt, Esiobu, Choudhary, Mahajan, Garcia-Olano, Perino, Hupkes, Lakomkin, AlBadawy, Lobanova, Dinan, Smith, Radenovic, Guzmán, Zhang, Synnaeve, Lee, Anderson, Thattai, Nail, Mialon, Pang, Cucurell, Nguyen, Korevaar, Xu, Touvron, Zarov, Ibarra, Kloumann, Misra, Evtimov, Zhang, Copet, Lee, Geffert, Vranes, Park, Mahadeokar, Shah, van~der Linde, Billock, Hong, Lee, Fu, Chi, Huang, Liu, Wang, Yu, Bitton, Spisak, Park, Rocca, Johnstun, Saxe, Jia, Alwala, Prasad, Upasani, Plawiak, Li, Heafield, Stone, El-Arini, Iyer, Malik, Chiu, Bhalla, Lakhotia, Rantala-Yeary, van~der Maaten, Chen, Tan, Jenkins, Martin, Madaan, Malo, Blecher,
  Landzaat, de~Oliveira, Muzzi, Pasupuleti, Singh, Paluri, Kardas, Tsimpoukelli, Oldham, Rita, Pavlova, Kambadur, Lewis, Si, Singh, Hassan, Goyal, Torabi, Bashlykov, Bogoychev, Chatterji, Zhang, Duchenne, Çelebi, Alrassy, Zhang, Li, Vasic, Weng, Bhargava, Dubal, Krishnan, Koura, Xu, He, Dong, Srinivasan, Ganapathy, Calderer, Cabral, Stojnic, Raileanu, Maheswari, Girdhar, Patel, Sauvestre, Polidoro, Sumbaly, Taylor, Silva, Hou, Wang, Hosseini, Chennabasappa, Singh, Bell, Kim, Edunov, Nie, Narang, Raparthy, Shen, Wan, Bhosale, Zhang, Vandenhende, Batra, Whitman, Sootla, Collot, Gururangan, Borodinsky, Herman, Fowler, Sheasha, Georgiou, Scialom, Speckbacher, Mihaylov, Xiao, Karn, Goswami, Gupta, Ramanathan, Kerkez, Gonguet, Do, Vogeti, Albiero, Petrovic, Chu, Xiong, Fu, Meers, Martinet, Wang, Wang, Tan, Xia, Xie, Jia, Wang, Goldschlag, Gaur, Babaei, Wen, Song, Zhang, Li, Mao, Coudert, Yan, Chen, Papakipos, Singh, Srivastava, Jain, Kelsey, Shajnfeld, Gangidi, Victoria, Goldstand, Menon, Sharma, Boesenberg,
  Baevski, Feinstein, Kallet, Sangani, Teo, Yunus, Lupu, Alvarado, Caples, Gu, Ho, Poulton, Ryan, Ramchandani, Dong, Franco, Goyal, Saraf, Chowdhury, Gabriel, Bharambe, Eisenman, Yazdan, James, Maurer, Leonhardi, Huang, Loyd, Paola, Paranjape, Liu, Wu, Ni, Hancock, Wasti, Spence, Stojkovic, Gamido, Montalvo, Parker, Burton, Mejia, Liu, Wang, Kim, Zhou, Hu, Chu, Cai, Tindal, Feichtenhofer, Gao, Civin, Beaty, Kreymer, Li, Adkins, Xu, Testuggine, David, Parikh, Liskovich, Foss, Wang, Le, Holland, Dowling, Jamil, Montgomery, Presani, Hahn, Wood, Le, Brinkman, Arcaute, Dunbar, Smothers, Sun, Kreuk, Tian, Kokkinos, Ozgenel, Caggioni, Kanayet, Seide, Florez, Schwarz, Badeer, Swee, Halpern, Herman, Sizov, Guangyi, Zhang, Lakshminarayanan, Inan, Shojanazeri, Zou, Wang, Zha, Habeeb, Rudolph, Suk, Aspegren, Goldman, Zhan, Damlaj, Molybog, Tufanov, Leontiadis, Veliche, Gat, Weissman, Geboski, Kohli, Lam, Asher, Gaya, Marcus, Tang, Chan, Zhen, Reizenstein, Teboul, Zhong, Jin, Yang, Cummings, Carvill, Shepard, McPhie,
  Torres, Ginsburg, Wang, Wu, U, Saxena, Khandelwal, Zand, Matosich, Veeraraghavan, Michelena, Li, Jagadeesh, Huang, Chawla, Huang, Chen, Garg, A, Silva, Bell, Zhang, Guo, Yu, Moshkovich, Wehrstedt, Khabsa, Avalani, Bhatt, Mankus, Hasson, Lennie, Reso, Groshev, Naumov, Lathi, Keneally, Liu, Seltzer, Valko, Restrepo, Patel, Vyatskov, Samvelyan, Clark, Macey, Wang, Hermoso, Metanat, Rastegari, Bansal, Santhanam, Parks, White, Bawa, Singhal, Egebo, Usunier, Mehta, Laptev, Dong, Cheng, Chernoguz, Hart, Salpekar, Kalinli, Kent, Parekh, Saab, Balaji, Rittner, Bontrager, Roux, Dollar, Zvyagina, Ratanchandani, Yuvraj, Liang, Alao, Rodriguez, Ayub, Murthy, Nayani, Mitra, Parthasarathy, Li, Hogan, Battey, Wang, Howes, Rinott, Mehta, Siby, Bondu, Datta, Chugh, Hunt, Dhillon, Sidorov, Pan, Mahajan, Verma, Yamamoto, Ramaswamy, Lindsay, Lindsay, Feng, Lin, Zha, Patil, Shankar, Zhang, Zhang, Wang, Agarwal, Sajuyigbe, Chintala, Max, Chen, Kehoe, Satterfield, Govindaprasad, Gupta, Deng, Cho, Virk, Subramanian, Choudhury,
  Goldman, Remez, Glaser, Best, Koehler, Robinson, Li, Zhang, Matthews, Chou, Shaked, Vontimitta, Ajayi, Montanez, Mohan, Kumar, Mangla, Ionescu, Poenaru, Mihailescu, Ivanov, Li, Wang, Jiang, Bouaziz, Constable, Tang, Wu, Wang, Wu, Gao, Kleinman, Chen, Hu, Jia, Qi, Li, Zhang, Zhang, Adi, Nam, Yu, Wang, Zhao, Hao, Qian, Li, He, Rait, DeVito, Rosnbrick, Wen, Yang, Zhao, and Ma}]{grattafiori2024llama3herdmodels}
Aaron Grattafiori, Abhimanyu Dubey, Abhinav Jauhri, Abhinav Pandey, Abhishek Kadian, Ahmad Al-Dahle, Aiesha Letman, Akhil Mathur, Alan Schelten, Alex Vaughan, Amy Yang, Angela Fan, Anirudh Goyal, Anthony Hartshorn, Aobo Yang, Archi Mitra, Archie Sravankumar, Artem Korenev, Arthur Hinsvark, and 542 others. 2024.
\newblock \href {https://arxiv.org/abs/2407.21783} {The llama 3 herd of models}.
\newblock \emph{Preprint}, arXiv:2407.21783.

\bibitem[{Hu et~al.(2024)Hu, Luo, and Feng}]{hu-etal-2024-ella}
Yutong Hu, Kangcheng Luo, and Yansong Feng. 2024.
\newblock \href {https://doi.org/10.18653/v1/2024.acl-demos.36} {{ELLA}: Empowering {LLM}s for interpretable, accurate and informative legal advice}.
\newblock In \emph{Proceedings of the 62nd Annual Meeting of the Association for Computational Linguistics (Volume 3: System Demonstrations)}, pages 374--387, Bangkok, Thailand. Association for Computational Linguistics.

\bibitem[{Lai et~al.(2023)Lai, Gan, Wu, Qi, and Yu}]{lai2023largelanguagemodelslaw}
Jinqi Lai, Wensheng Gan, Jiayang Wu, Zhenlian Qi, and Philip~S. Yu. 2023.
\newblock \href {https://arxiv.org/abs/2312.03718} {Large language models in law: A survey}.
\newblock \emph{Preprint}, arXiv:2312.03718.

\bibitem[{{LangChain}(2026)}]{langgraph}
{LangChain}. 2026.
\newblock Langgraph: Agent orchestration framework for reliable ai agents.
\newblock \url{https://www.langchain.com/langgraph}.
\newblock Accessed: 2026-07-11.

\bibitem[{Lewis et~al.(2020)Lewis, Perez, Piktus, Petroni, Karpukhin, Goyal, K\"{u}ttler, Lewis, Yih, Rockt\"{a}schel, Riedel, and Kiela}]{NEURIPS2020_6b493230}
Patrick Lewis, Ethan Perez, Aleksandra Piktus, Fabio Petroni, Vladimir Karpukhin, Naman Goyal, Heinrich K\"{u}ttler, Mike Lewis, Wen-tau Yih, Tim Rockt\"{a}schel, Sebastian Riedel, and Douwe Kiela. 2020.
\newblock \href {https://proceedings.neurips.cc/paper_files/paper/2020/file/6b493230205f780e1bc26945df7481e5-Paper.pdf} {Retrieval-augmented generation for knowledge-intensive nlp tasks}.
\newblock In \emph{Advances in Neural Information Processing Systems}, volume~33, pages 9459--9474. Curran Associates, Inc.

\bibitem[{Li et~al.(2025)Li, Wu, Liu, Cai, Qing, Wang, Kang, Liu, Wu, and Kuang}]{li-etal-2025-unilr}
Ang Li, Yiquan Wu, Yifei Liu, Ming Cai, Lizhi Qing, Shihang Wang, Yangyang Kang, Chengyuan Liu, Fei Wu, and Kun Kuang. 2025.
\newblock \href {https://doi.org/10.18653/v1/2025.acl-long.584} {{U}ni{LR}: Unleashing the power of {LLM}s on multiple legal tasks with a unified legal retriever}.
\newblock In \emph{Proceedings of the 63rd Annual Meeting of the Association for Computational Linguistics (Volume 1: Long Papers)}, pages 11953--11967, Vienna, Austria. Association for Computational Linguistics.

\bibitem[{Li et~al.(2024)Li, Wang, Ding, and Chen}]{li2024largelanguagemodelsfinance}
Yinheng Li, Shaofei Wang, Han Ding, and Hang Chen. 2024.
\newblock \href {https://arxiv.org/abs/2311.10723} {Large language models in finance: A survey}.
\newblock \emph{Preprint}, arXiv:2311.10723.

\bibitem[{Luo et~al.(2025)Luo, Huang, Jiang, and Feng}]{luo-etal-2025-automating}
Kangcheng Luo, Quzhe Huang, Cong Jiang, and Yansong Feng. 2025.
\newblock \href {https://doi.org/10.18653/v1/2025.acl-long.204} {Automating legal interpretation with {LLM}s: Retrieval, generation, and evaluation}.
\newblock In \emph{Proceedings of the 63rd Annual Meeting of the Association for Computational Linguistics (Volume 1: Long Papers)}, pages 4015--4047, Vienna, Austria. Association for Computational Linguistics.

\bibitem[{{OpenAI}(2026)}]{openai2026gpt54}
{OpenAI}. 2026.
\newblock Gpt-5.4.
\newblock \url{https://openai.com/index/introducing-gpt-5-4/}.
\newblock Accessed: 2026-07-10.

\bibitem[{OpenAI et~al.(2024)OpenAI, Achiam, Adler, Agarwal, Ahmad, Akkaya, Aleman, Almeida, Altenschmidt, Altman, Anadkat, Avila, Babuschkin, Balaji, Balcom, Baltescu, Bao, Bavarian, Belgum, Bello, Berdine, Bernadett-Shapiro, Berner, Bogdonoff, Boiko, Boyd, Brakman, Brockman, Brooks, Brundage, Button, Cai, Campbell, Cann, Carey, Carlson, Carmichael, Chan, Chang, Chantzis, Chen, Chen, Chen, Chen, Chen, Chess, Cho, Chu, Chung, Cummings, Currier, Dai, Decareaux, Degry, Deutsch, Deville, Dhar, Dohan, Dowling, Dunning, Ecoffet, Eleti, Eloundou, Farhi, Fedus, Felix, Fishman, Forte, Fulford, Gao, Georges, Gibson, Goel, Gogineni, Goh, Gontijo-Lopes, Gordon, Grafstein, Gray, Greene, Gross, Gu, Guo, Hallacy, Han, Harris, He, Heaton, Heidecke, Hesse, Hickey, Hickey, Hoeschele, Houghton, Hsu, Hu, Hu, Huizinga, Jain, Jain, Jang, Jiang, Jiang, Jin, Jin, Jomoto, Jonn, Jun, Kaftan, Łukasz Kaiser, Kamali, Kanitscheider, Keskar, Khan, Kilpatrick, Kim, Kim, Kim, Kirchner, Kiros, Knight, Kokotajlo, Łukasz Kondraciuk,
  Kondrich, Konstantinidis, Kosic, Krueger, Kuo, Lampe, Lan, Lee, Leike, Leung, Levy, Li, Lim, Lin, Lin, Litwin, Lopez, Lowe, Lue, Makanju, Malfacini, Manning, Markov, Markovski, Martin, Mayer, Mayne, McGrew, McKinney, McLeavey, McMillan, McNeil, Medina, Mehta, Menick, Metz, Mishchenko, Mishkin, Monaco, Morikawa, Mossing, Mu, Murati, Murk, Mély, Nair, Nakano, Nayak, Neelakantan, Ngo, Noh, Ouyang, O'Keefe, Pachocki, Paino, Palermo, Pantuliano, Parascandolo, Parish, Parparita, Passos, Pavlov, Peng, Perelman, de~Avila Belbute~Peres, Petrov, de~Oliveira~Pinto, Michael, Pokorny, Pokrass, Pong, Powell, Power, Power, Proehl, Puri, Radford, Rae, Ramesh, Raymond, Real, Rimbach, Ross, Rotsted, Roussez, Ryder, Saltarelli, Sanders, Santurkar, Sastry, Schmidt, Schnurr, Schulman, Selsam, Sheppard, Sherbakov, Shieh, Shoker, Shyam, Sidor, Sigler, Simens, Sitkin, Slama, Sohl, Sokolowsky, Song, Staudacher, Such, Summers, Sutskever, Tang, Tezak, Thompson, Tillet, Tootoonchian, Tseng, Tuggle, Turley, Tworek, Uribe, Vallone,
  Vijayvergiya, Voss, Wainwright, Wang, Wang, Wang, Ward, Wei, Weinmann, Welihinda, Welinder, Weng, Weng, Wiethoff, Willner, Winter, Wolrich, Wong, Workman, Wu, Wu, Wu, Xiao, Xu, Yoo, Yu, Yuan, Zaremba, Zellers, Zhang, Zhang, Zhao, Zheng, Zhuang, Zhuk, and Zoph}]{openai2024gpt4technicalreport}
OpenAI, Josh Achiam, Steven Adler, Sandhini Agarwal, Lama Ahmad, Ilge Akkaya, Florencia~Leoni Aleman, Diogo Almeida, Janko Altenschmidt, Sam Altman, Shyamal Anadkat, Red Avila, Igor Babuschkin, Suchir Balaji, Valerie Balcom, Paul Baltescu, Haiming Bao, Mohammad Bavarian, Jeff Belgum, and 262 others. 2024.
\newblock \href {https://arxiv.org/abs/2303.08774} {Gpt-4 technical report}.
\newblock \emph{Preprint}, arXiv:2303.08774.

\bibitem[{Panchal et~al.(2025)Panchal, Gole, Narute, and Joshi}]{panchal2025lawpalretrievalaugmented}
Dnyanesh Panchal, Aaryan Gole, Vaibhav Narute, and Raunak Joshi. 2025.
\newblock \href {https://arxiv.org/abs/2502.16573} {Lawpal : A retrieval augmented generation based system for enhanced legal accessibility in india}.
\newblock \emph{Preprint}, arXiv:2502.16573.

\bibitem[{T.y.s.s et~al.(2025)T.y.s.s, Elkhayat, Ichim, Shetty, Wang, Ma, Nourbakhsh, and Liu}]{t-y-s-s-etal-2025-cocolex}
Santosh T.y.s.s, Youssef~Tarek Elkhayat, Oana Ichim, Pranav Shetty, Dongsheng Wang, Zhiqiang Ma, Armineh Nourbakhsh, and Xiaomo Liu. 2025.
\newblock \href {https://doi.org/10.18653/v1/2025.acl-long.931} {{C}o{C}o{L}ex: Confidence-guided copy-based decoding for grounded legal text generation}.
\newblock In \emph{Proceedings of the 63rd Annual Meeting of the Association for Computational Linguistics (Volume 1: Long Papers)}, pages 19002--19018, Vienna, Austria. Association for Computational Linguistics.

\bibitem[{Wei et~al.(2025)Wei, Shi, Tong, Reddy, Natarajan, Jain, Garimella, and Huang}]{wei-etal-2025-legalcore}
Kangda Wei, Xi~Shi, Jonathan Tong, Sai~Ramana Reddy, Anandhavelu Natarajan, Rajiv Jain, Aparna Garimella, and Ruihong Huang. 2025.
\newblock \href {https://doi.org/10.18653/v1/2025.findings-acl.1284} {{L}egal{C}ore: A dataset for event coreference resolution in legal documents}.
\newblock In \emph{Findings of the Association for Computational Linguistics: ACL 2025}, pages 25044--25059, Vienna, Austria. Association for Computational Linguistics.

\bibitem[{Yue et~al.(2025)Yue, Huang, Jia, Wang, Liu, Song, Huang, and Wei}]{shengbinyue-etal-2025-multi}
Shengbin Yue, Ting Huang, Zheng Jia, Siyuan Wang, Shujun Liu, Yun Song, Xuanjing Huang, and Zhongyu Wei. 2025.
\newblock \href {https://doi.org/10.18653/v1/2025.findings-naacl.365} {Multi-agent simulator drives language models for legal intensive interaction}.
\newblock In \emph{Findings of the Association for Computational Linguistics: NAACL 2025}, pages 6552--6585, Albuquerque, New Mexico. Association for Computational Linguistics.

\end{thebibliography}
\onecolumn
\appendix

\begingroup
\setlist[itemize]{leftmargin=*, itemsep=0pt, topsep=0pt, parsep=0pt, partopsep=0pt}

\section{Case Study of Deep Research}
\label{sec:abridged_legal_qa_example}

{\normalsize
\setlength{\tabcolsep}{3pt}
\renewcommand{\arraystretch}{1.0}
\setlength{\extrarowheight}{0pt}
\setlength{\LTpre}{0pt}
\setlength{\LTpost}{0pt}
\setlength{\parskip}{0pt}

\begin{longtable}{@{}
>{\raggedright\arraybackslash}p{0.492\linewidth}|
>{\raggedright\arraybackslash}p{0.492\linewidth}
@{}}

\endfirsthead
\hline
\endhead
\endfoot
\endlastfoot

\hline
\textbf{$Q$: 被辞退可以赔多少钱？}
&
\textbf{$Q$: How Much Compensation Can I Receive If I Am Dismissed?}
\\
\hline

\textbf{$R$: 1. 问题概述}

用户咨询“被辞退可以赔多少钱”。该问题通常取决于辞退性质：

\begin{itemize}
    \item 合法解除且依法应补偿：一般为经济补偿金，即通常所称 $N$；
    \item 特定无过失性解除且未提前30日书面通知：可能为 $N+1$；
    \item 违法解除/终止：劳动者可选择要求继续履行劳动合同，或在不能/不要求继续履行时主张违法解除赔偿金，通常按经济补偿标准的二倍，即 $2N$；
    \item 高收入员工：经济补偿基数可能受“当地上年度职工月平均工资三倍”和“最高12年”限制，具体以用人单位所在地统计口径为准[网1][网2]。
\end{itemize}
&
\textbf{$R$: 1. Overview and Issue Identification}

The user asks, “How much compensation can I receive if I am dismissed?” The answer generally depends on the legal basis and nature of the dismissal:

\begin{itemize}
    \item Lawful termination for which financial compensation is required by law: the employee will generally be entitled to financial compensation, commonly expressed as “$N$.”
    \item Certain employer-initiated terminations on statutory non-fault grounds without 30 days' prior written notice: the employee may be entitled to “$N+1$.”
    \item Unlawful revocation or termination of the labor contract: the employee may choose to request continued performance of the labor contract or, where continued performance is impossible or not requested, claim compensation for unlawful termination, generally equal to twice the statutory financial-compensation amount, or “$2N$.”
    \item High-income employees: the wage base for financial compensation may be subject to a cap of three times the local average monthly wage for employees in the preceding year and a maximum compensable service period of 12 years. The applicable statistical measure depends on the locality in which the employing unit is located [Web 1][Web 2].
\end{itemize}
\\

\textbf{2. 法律分析}

\textbf{结论一：被辞退能拿多少，首先看解除是否合法、属于哪类解除}
&
\textbf{2. Detailed Legal Analysis}

\textbf{Conclusion 1: The amount recoverable first depends on whether the termination was lawful and what type of termination occurred}
\\

\textbf{结论二：一般经济补偿金“$N$”的计算口径是“工作年限 $\times$ 月工资”}
&
\textbf{Conclusion 2: The general formula for “$N$” financial compensation is “years of service $\times$ monthly wage”} 
\\

\textbf{结论三：“$N+1$”不是所有辞退都有，只适用于特定无过失性解除场景}
&
\textbf{Conclusion 3: “$N+1$” does not apply to every dismissal; it is limited to certain terminations on statutory non-fault grounds}
\\

\textbf{结论四：违法辞退通常可主张“继续履行”或“$2N$赔偿金”}
&
\textbf{Conclusion 4: In cases of unlawful dismissal, the employee may generally seek continued performance or “$2N$” compensation}
\\

\textbf{结论五：高收入员工可能受“三倍社平工资 + 最多12年”封顶限制}
&
\textbf{Conclusion 5: High-income employees may be subject to the “three-times-local-average-wage plus 12-year maximum” cap}
\\

\textbf{3. 相关法条}

按效力位阶及指定顺序列示：法律 $>$ 行政法规。

\textbf{法律}

\textbf{《中华人民共和国劳动合同法》第87条}

用人单位违法解除或终止劳动合同的，通常按经济补偿标准二倍支付赔偿金。

&
\textbf{3. Statutory Provisions}

The authorities are listed by level of legal effect and in the prescribed order: laws $>$ administrative regulations. 

\textbf{Laws}

\textbf{Article 87 of the Labor Contract Law of the People's Republic of China}

Where an employing unit unlawfully revokes or terminates a labor contract, it is generally required to pay compensation at twice the statutory financial-compensation rate.

\\

\textbf{4. 案例参考}

现有案例均为“普通案例”，以下保留一个代表性案例。

\textbf{云南某某有限公司与陈某经济补偿金纠纷案，案号：（2024）云25民终1139号}

要旨：用人单位主张无需支付经济补偿金26001.87元，反映解除或终止劳动关系后经济补偿支付义务的常见争议[案12]。
&
\textbf{4. Case References}

All of the available cases are classified as “ordinary cases.” One representative case is retained below.

\textbf{Dispute over financial compensation between Yunnan [Redacted] Co., Ltd. and Chen [given name withheld], Case No. (2024)云25民终1139号}

Key point: The employing unit argued that it was not required to pay RMB 26,001.87 in financial compensation. The case reflects a common dispute over the obligation to pay financial compensation after the revocation or termination of an employment relationship [Case 12].
\\

\textbf{5. 风险提示}

\textbf{不要只看“被辞退”三个字，要先判断解除类型}

主动辞职、协商解除、合法过失性解除、无过失性解除、经济性裁员、违法解除，对应金额差异很大。
&
\textbf{5. Risk Alerts}

\textbf{Do not focus only on the fact that the employee was “dismissed”; first identify the legal category of termination}

Voluntary resignation, termination by mutual agreement, lawful termination for employee fault, termination on statutory non-fault grounds, economic redundancy, and unlawful termination may produce substantially different payment outcomes.
\\

\textbf{6. 免责声明}\quad

本报告仅根据已提供的研究结果和可引用来源进行一般性法律分析，仅供参考，不构成正式法律意见。
&
\textbf{6. Liability Disclaimer}\quad

This report provides only a general legal analysis based on the research results and citable sources supplied. It is for reference purposes only and does not constitute formal legal advice.
\\

\hline
\caption{An abridged bilingual Deep Research report; detailed analyses, additional legal materials, and case references are omitted for space.}
\label{tab:appendix_abridged_deep_research}
\\

\end{longtable}
}

\endgroup

\end{CJK*}
\end{document}